\documentclass[letterpaper, 10 pt, conference]{ieeeconf}  

\IEEEoverridecommandlockouts                              

\usepackage{graphicx}
\usepackage{algorithm}
\usepackage{algpseudocode}
\usepackage{amsmath}

\usepackage{caption}
\usepackage{color}
\usepackage{subcaption}
\usepackage{tikz}
\usepackage{booktabs}

\title{\LARGE \bf
 Robotic Control Optimization Through Kernel Selection in Safe Bayesian Optimization
}

\author{\thanks{This work was supported by RIE2025 Manufacturing, Trade and Connectivity (MTC) Industry Alignment Fund – Pre-Positioning (IAF-PP) under Grant M22K4a0044 through WP3-Energy Efficient Motor Drive System with GaN-based Traction Inverters. (\textit{Corresponding author: Xiaocong Li.})} Lihao Zheng\thanks{Lihao Zheng is with the School of Data Science, The Chinese University of Hong Kong, Shenzhen, China. This work was done while he was a Research Intern at the Singapore Institute of Manufacturing Technology, A*STAR. Email: \texttt{lihaozheng@link.cuhk.edu.cn}}, Hongxuan Wang\thanks{Hongxuan Wang and Prahlad Vadakkepat are with the Department of Electrical and Computer Engineering, National University of Singapore, Singapore 117583. Email: \texttt{hongxuanwang@u.nus.edu}, \texttt{prahlad@nus.edu.sg}}, Xiaocong Li\thanks{Xiaocong Li is with the Singapore Institute of Manufacturing Technology, Agency for Science, Technology and Research (A*STAR), Singapore 138634. Email: \texttt{li\_xiaocong@simtech.a-star.edu.sg}}, Jun Ma\thanks{Jun Ma is with the Robotics and Autonomous Systems Thrust, The Hong Kong University of Science and Technology (Guangzhou), Guangzhou, China. Email: \texttt{jun.ma@ust.hk}}, and Prahlad Vadakkepat 
}

\begin{document}

\maketitle
\thispagestyle{empty}
\pagestyle{empty}

\begin{abstract}

Control system optimization has long been a fundamental challenge in robotics. While recent advancements have led to the development of control algorithms that leverage learning-based approaches, such as SafeOpt, to optimize single feedback controllers, scaling these methods to high-dimensional complex systems with multiple controllers remains an open problem. In this paper, we propose a novel learning-based control optimization method, which enhances the additive Gaussian process-based Safe Bayesian Optimization algorithm to efficiently tackle high-dimensional problems through kernel selection. We use PID controller optimization in drones as a representative example and test the method on Safe Control Gym, a benchmark designed for evaluating safe control techniques. We show that the proposed method provides a more efficient and optimal solution for high-dimensional control optimization problems, demonstrating significant improvements over existing techniques.

\end{abstract}

\section{Introduction}

To address the widely adopted yet time-consuming and often suboptimal process of manual controller tuning, the demand for automatic controller optimization has grown across various fields, including robotics \cite{ROVEDA2020104488, Huang21RAL}, automotive systems \cite{automotive}, and industrial automation \cite{Li24TMech}. Most previous methods are based on a simplified system model and determine the optimal controller parameters through the system model \cite{aastrom1984automatic, hang2002relay}. Such methods are often prone to suboptimal results due to a less accurate system model and the influence of the noise \cite{Berkenkamp16}. With the development of data-driven methods, optimizing controller parameters based on real robotic system motion data has shown notable improvements in control performance, such as methods using Iterative Feedback Tuning (IFT) \cite{IFT}, Variable gain control \cite{variable_gain}, and Bayesian optimization \cite{mockus2005bayesian, Khosravi22TIE}. 

Bayesian optimization is often used to optimize black-box objective functions that are expensive to evaluate. It employs a surrogate model, typically a Gaussian Process (GP), to iteratively predict the function’s distribution and guide the search for the optimum using an acquisition function. Since the noise measurement can be modeled as a GP together with the objective function \cite{Srinivas_2012}, Bayesian optimization is particularly effective for tasks like optimizing PID controller parameters in control systems. However, Bayesian optimization often evaluates unsafe parameters in practical applications \cite{Berkenkamp16} since the iterative process does not consider the physical meaning of the evaluated function. For example, unsafe parameter combinations in quadrotor control could lead to collisions with walls or ceilings. Besides, Bayesian optimization uses Gaussian kernels for GP calculations, which is prone to the curse of dimensionality \cite{wang24additiveBO}. When the dimension of the problem becomes larger than 3, the number of evaluations required by Bayesian optimization will increase significantly. Hence, controller parameter tuning of high-order complex systems remains a significant challenge, especially when safety and robustness should be guaranteed \cite{Jiang24TAI, Khosravi23TIE}.

\begin{figure}[t]
    \centering
    \setlength{\belowcaptionskip}{-18pt}
    \includegraphics[width=\linewidth]{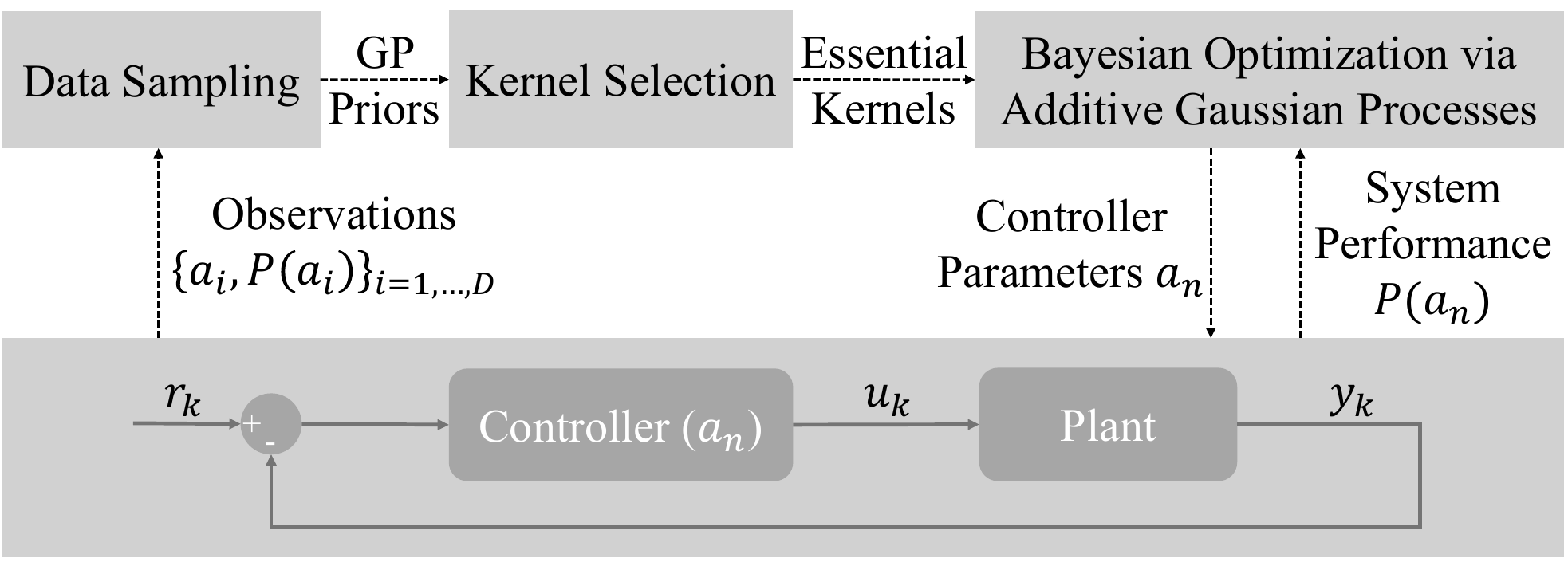}
    \caption{\small{Overview of the method. First, the input and output of the system are measured to obtain a set of observations, which serve as the GP priors for calculating kernel selection. Subsequently, Bayesian optimization via additive Gaussian processes uses the most important one or several additive kernels calculated by kernel selection for optimization, iteratively selecting safe new parameters and testing the performance in the system until the optimal control parameters are obtained.}}
    \label{fig:KSflow}
\end{figure}

Following this line of research, Sui et al. \cite{Sui15} proposed SafeOpt. This first safety-aware Bayesian optimization algorithm ensures the safety of the optimization process by avoiding evaluating parameters whose performance values fall below the pre-defined safe threshold. However, it is less feasible in high dimensions. Kirschner et al. \cite{linebo} then proposed the LINEBO algorithm to address the challenges of optimizing high-dimensional problems by breaking them into manageable one-dimensional sub-problems. Still, it is hard to apply to control engineering because it usually takes hundreds of evaluations, which may cause severe wear to the system. Wang et al. \cite{wang24additiveBO} proposed the high-dimensional safe Bayesian optimization algorithm via additive Gaussian processes, which is more efficient for control optimization. They employed an additive structure to the traditional Gaussian kernels to enhance the information acquisition efficiency, but it brought too much complexity to the calculation. When the problem dimension gets higher than 6, the experimental validation with actual hardware will be hard to implement.

Dimensionality reduction based on feature selection is a standard method to reduce the calculation for complex algorithms \cite{pca, wang2016ae, wang2024gradient}. In this framework, the kernel-target alignment method \cite{KernelAlignment} measures the similarity between kernels and the objective function, providing a way to select essential kernels based on the degree of agreement with the learning task. While this approach offers theoretical insights and can improve model accuracy, it is computationally intensive and inefficient in high dimensions. To address these limitations, Ding et al. \cite{KernelSelection} introduce an alternative approach using the Nyström kernel matrix approximation. This method is more computationally efficient and provides theoretical consistency guarantees, making it a better choice for real-time applications. 

In this paper, we propose to adopt the Nyström approximation method to ensure efficient and effective kernel selection to identify the most important additive kernels based on the GP prior, and then use the selected kernels to optimize high-dimensional control problems safely and efficiently. The process is illustrated in Fig. \ref{fig:KSflow}. We validated our algorithm using the general benchmark Safe Control Gym \cite{safecontrolgym}, and experimental results demonstrate that our method outperforms existing high-dimensional safe Bayesian optimization algorithms for quadrotor trajectory control.

\section{Problem Statement}

This paper considers the safe optimization problem of high-dimensional controllers. In this section, we will first introduce the definition of safe control optimization, then introduce the high-dimensional benchmark environment and criteria for methods evaluation.

\subsection{Safe Control Optimization}

Assume we have a nonlinear control law as in Fig. \ref{fig:KSflow}:
\begin{equation}
    u_k = f(a, r_k, y_k),
\end{equation}
where $u_k$ is the control signal at time step $k$, $a \in \mathbf{A}$ is the controller parameters, $r_k$ is the reference signal, and $y_k$ is the plant output. We can define a performance measure $P$ that reflects the control objective based on $y_k$ and $r_k$ (e.g., tracking error), as well as some other internal signals like current and voltage. Control optimization aims to obtain the controller parameter $a^*$ that can maximize $P$ through iterations, so we model $P$ as a function of $a_n$ at iteration $n$, $P(a_n): \mathbf{A} \to \mathcal{R}$, which can be evaluated on the system. 

With an assumption that a default controller and its performance $(a_0, P(a_0))$ is known, we then have to solve the sequential optimization problem by selecting $a_1, a_2, \dots, a_n \in \mathbf{A}$, and get the performance evaluation after each iteration, until the performance converges to an optimum. In control engineering, the performance objective is usually to get a low steady-state tracking error while keeping a fast transient response and less overshoot.

When doing the above process, we assume that each step of the system iteration must be safe to avoid the evaluation of unstable controllers, which may bring excessive system wear and even damage. Our control optimization achieves safety by setting a minimum performance threshold $P_{min}$, that is, avoiding evaluating controller parameters whose performance may fall below the minimum threshold. By adding such constraints to the performance function, we can ensure the safety of the optimization process with a high probability.

\begin{figure}[t]
      \centering
      \setlength{\belowcaptionskip}{-15pt}
      \includegraphics[scale=0.3]{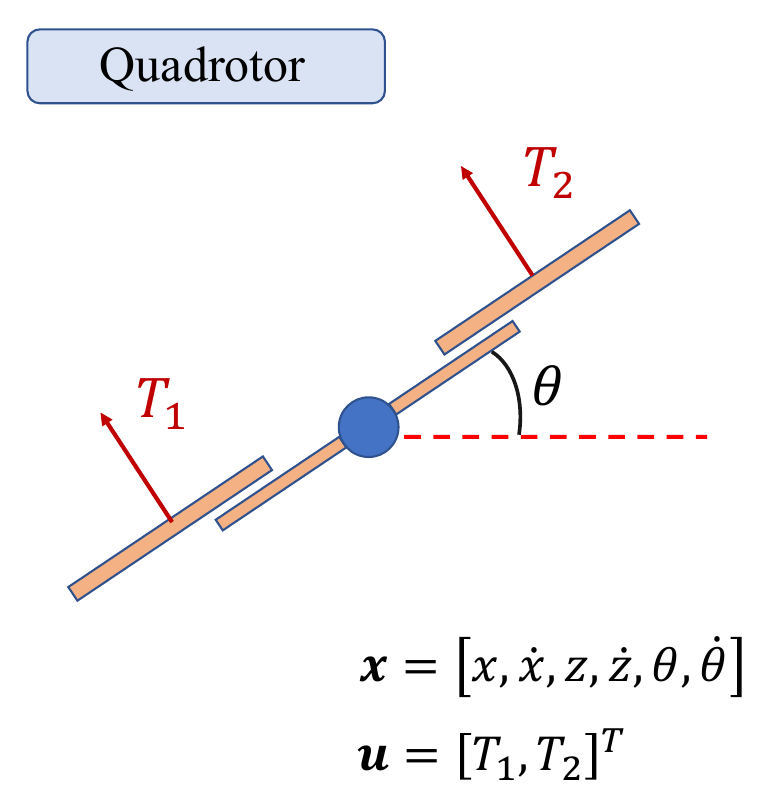}
      \caption{\small{Quadrotor model used for benchmark environment.}}
      \label{fig:quadrotor}
\end{figure}

\subsection{Benchmark Environment}

We evaluate the proposed method and compare it with baseline methods using Safe Control Gym \cite{safecontrolgym}. It offers a comprehensive benchmark suite of simulation environments to facilitate the development and evaluation of safe learning-based control algorithms. This suite includes a variety of control algorithms, ranging from traditional methods like the Linear Quadratic Regulator (LQR) and Iterative LQR (iLQR) \cite{buchli2017optimal}, to modern reinforcement learning approaches such as Proximal Policy Optimization (PPO) \cite{PPO} and Soft Actor-Critic (SAC) \cite{SAC}. We use $2D$ quadrotor control for trajectory tracking in Safe Control Gym to show how our method optimizes control tasks effectively and safely. We demonstrate using PID control due to its widespread use in robotics and industrial applications, particularly in quadrotor control.

The quadrotor motion is constrained to the $2D$ $xz$-plane. As shown in Fig. \ref{fig:quadrotor}, the states of the quadrotor are defined by the vector \( \textbf{x} = [x, \dot{x}, z, \dot{z}, \theta, \dot{\theta}]^T \), where \( x \) and \( z \) represent the horizontal and vertical positions, and \( \dot{x} \) and \( \dot{z} \) are the corresponding velocities. The pitch angle \( \theta \) and angular velocity \( \dot{\theta} \) describe the orientation and rotational dynamics of the quadrotor. The system is controlled by the thrusts \( \textbf{u} = [T_1, T_2]^T \), generated by two pairs of motors positioned on either side of the body’s $y$-axis. The dynamics of the quadrotor are governed by the accelerations:

\begin{equation}
\ddot{x} = \sin \theta \frac{(T_1 + T_2)}{m},
\end{equation}

\begin{equation}
\ddot{z} = \cos \theta \frac{(T_1 + T_2)}{m} - g,
\end{equation}

\begin{equation}
\ddot{\theta} = \frac{(T_2 - T_1) d}{I_{yy}},
\end{equation}
where \( m \) is the mass of the quadrotor, \( g \) is the acceleration due to gravity, \( d = l / \sqrt{2} \) is the effective moment arm (with \( l \) being the arm length of the quadrotor), and \( I_{yy} \) is the moment of inertia about the y-axis.

We fix the parameters of the speed controllers and achieve optimal trajectory tracking by tuning the PID parameters of the three position controllers, denoted as $\tilde{\textbf{x}} = [x, z, \theta]^T$. In our task, we focused on ensuring the quadrotor’s position closely follows the reference trajectory by adjusting the PID parameters specifically for position control. Since each PID controller consists of three parameters—$P$, $I$, and $D$—and we have three position controllers, this results in a total of 9 parameters (3 for each controller) to be tuned, making the problem 9-dimensional.

\subsection{Evaluation Criteria}

As our goal is to obtain the optimal controller parameters that maximize the performance $P(a_n)$ during the trajectory tracking, we design the quadratic cost at iteration $n$ as:
\begin{equation}
\begin{aligned}
J^Q &= \frac{1}{2} \sum_{i=0}^{L} (\tilde{\textbf{x}}_i - \tilde{\textbf{x}}_i^{\text{ref}})^T \mathbf{Q} (\tilde{\textbf{x}}_i - \tilde{\textbf{x}}_i^{\text{ref}}) \\
&\quad + \frac{1}{2} \sum_{i=0}^{L-1} (\mathbf{u}_i - \mathbf{u}_i^{\text{ref}})^T \mathbf{R} (\mathbf{u}_i - \mathbf{u}_i^{\text{ref}}),
\end{aligned}
\end{equation}
where $\tilde{\textbf{x}}_i^{\text{ref}}$ and $\mathbf{u}_i^{\text{ref}}$ are the reference states and control signals for all \( i \in \{0, \ldots, L\} \), and \( L \) is the number of control steps in one iteration. Our performance measurement $P(a_n)$ is designed using this cost function \( J^Q \). As Bayesian optimization algorithms require positive performance rewards, and larger rewards represent better performances, we define the performance reward \( J^R = -J^Q \) and scale \(J^R\) to a positive number. 

We also consider the safety of the system in the design of performance measurement. In our task, \( L_{\text{expected}} \) is defined as the expected control steps of the entire iteration. When \( L < L_{\text{expected}} \), it is indicated that the trajectory tracking is not completed, and a common reason is that the quadrotor hits the boundary of the environment, which means the current controller is unsafe. Given the situation, our performance function $P(a_n)$ is designed to combine the scaled reward \(J^R_{\text{scaled}}\) and \(L\) as follows:

\begin{equation}
\begin{aligned}
P(a_n) = J^R_{\text{scaled}} - \left(1 - \lambda \left(\frac{L}{L_{\text{expected}}}\right)^2 \right),
\end{aligned}
\end{equation}
where \(\lambda\) is a variable to adapt the value of the performance $P(a_n)$. The second term in $P(a_n)$ is a penalty if the trajectory tracking is incomplete.

The design of our performance measurement combines trajectory tracking performance and controller safety, so it is well-suited as an evaluation criterion for controller parameter optimization. By defining $P_{min} = 0$ without loss of generality, our safety constraint is defined such that a negative performance function value indicates an unsafe condition.




\section{Methods}






We use kernel selection to improve the efficiency of the additive Gaussian process-based safe Bayesian optimization algorithm \cite{wang24additiveBO}. In this section, we will introduce the components of our proposed method.

\subsection{Overview of Additive Gaussian Processes} 

Additive Gaussian Processes proposed by Duvenaud et al. \cite{AdditiveGaussianProcesses} combines one-dimensional Gaussian kernels with a high-dimensional additive structure. The additive kernels for different orders are computed as follows:
\begin{equation}
\begin{aligned}
k_{\text{add}_1}(a, a') &= \sum_{i=1}^D z_i, \\
k_{\text{add}_2}(a, a') &= \sum_{i=1}^{D-1} \sum_{j=i+1}^D z_i z_j, \\
k_{\text{add}_n}(a, a') &= \sum_{1 \leq i_1 < i_2 < \ldots < i_n \leq D} \prod_{d=1}^N z_{i_d}, \label{eq13}
\end{aligned}
\end{equation}
where \( z_i \) represents the one-dimensional base kernel for the \( i \)-th dimension. By summing up the products of the base kernels across various dimensions, we create high-dimensional additive kernels that capture interactions between different dimensions. This structure allows for a more comprehensive exploration of the search space, ultimately leading to improved performance in high-dimensional optimization tasks.

\subsection{Overview of Safe Bayesian Optimization}

Safe Bayesian optimization leverages Gaussian processes (GP) to approximate unknown objective functions and optimize the function without violating safety constraints. By defining an appropriate covariance function \( k(x_i, x_j) \), GP can combine past observations to predict the mean and variance of the objective function at unobserved points:
\begin{equation}
\begin{aligned}
\mu_n(x) &= \mathbf{k}_n(x)(\mathbf{K}_n + I_n \sigma_\omega^2)^{-1} \tilde{\mathbf{J}}_n \\
\sigma_n^2(x) &= k(x, x) - \mathbf{k}_n(x)(\mathbf{K}_n + I_n \sigma_\omega^2)^{-1}\mathbf{k}_n^T(x),
\end{aligned}
\end{equation}
where:
\begin{itemize}
    \item \( \mu_n(x) \) is the predicted mean of the unknown function at point \( x \),
    \item \( \sigma_n^2(x) \) is the predicted variance,
    \item \( \mathbf{k}_n(x) \) is the covariance vector between \( x \) and the observed points,
    \item \( \mathbf{K}_n \) is the covariance matrix of the observed points,
    \item \( \tilde{\mathbf{J}}_n = [\tilde{J}(x_1), \dots, \tilde{J}(x_n)]^T \) is the vector of noisy performance measurements.
\end{itemize}

Using these mean and variance estimates, the upper and lower bounds of the confidence interval for the function value at each point \( x \) can be calculated as:
\begin{equation}
\begin{aligned}
u_n(x) &= \mu_{n-1}(x) + \beta_n \sigma_{n-1}(x) \\
l_n(x) &= \mu_{n-1}(x) - \beta_n \sigma_{n-1}(x),
\end{aligned}
\end{equation}
where \( \beta_n \) is a variable that defines the confidence interval. Previous safe Bayesian optimization algorithms \cite{wang24additiveBO, Sui15, linebo} use these upper and lower bounds to define a safe set \( S_n \), containing all parameters \( x \) that have a high probability of satisfying the safety constraints. Safe Bayesian optimization algorithms ensure safety by only selecting the next point within \( S_n \) using some acquisition functions like GP-UCB \cite{Srinivas10} and Expected Improvement (EI) \cite{EI}.





\begin{algorithm}[b]
\caption{Nyström Approximate Kernel Selection}
\label{algorithm}
\renewcommand{\algorithmicrequire}{\textbf{Input:}}
\renewcommand{\algorithmicensure}{\textbf{Output:}}
\begin{algorithmic}[1]
\Require $\mathcal{S} = \{(x_i, y_i)\}_{i=1}^l$, $\mathcal{K} = \{\kappa_1, \dots, \kappa_n\}$, $c$, $k$, $\mu$;
\Ensure A list of $C_{\text{rec}}$ values for each kernel in the set $\mathcal{K}$

\State \textbf{Initialize}: $C_{\text{ree}}\_\text{list} = []$
\For{each $\kappa \in \mathcal{K}$}
    \State Sample $c$ indices from $\{1, \dots, l\}$ to form the index set $\mathcal{I}$;
    \State Generate $\textbf{C}$ and $\textbf{W}$ using $\mathcal{S}$ and $k$ according to $\mathcal{I}$;
    \State Calculate the SVD of $\mathbf{W}$ as $\mathbf{W = U_W \Sigma_W U_W^{\top}}$;

    \State Let $\mathbf{V = CU_{W,k} \sqrt{\Sigma^{+}_{W,k}}}$;
    \State Solve $\left(\mu \mathbf{I_k} + \textbf{V}^{\top} \mathbf{V}\right) \mathbf{t} \mathbf{= V^{\top} y}$ to obtain $\mathbf{t}$;
    \State $\mathbf{u} = \frac{1}{\mu l}(\mathbf{y} - \mathbf{V}t)$;
    \State $C_{\text{ree}}(\mathbf{\tilde{K}}) = \mu \mathbf{y^{\top} u}$;
    \State Append $C_{\text{ree}}(\mathbf{\tilde{K}})$ to $C_{\text{ree}}\_\text{list}$
\EndFor
\State \Return $C_{\text{ree}}\_\text{list}$
\end{algorithmic}
\end{algorithm}

\subsection{Kernel Selection}
\label{method:kernel_selection}

As the search space of Gaussian kernels in high dimensions is very limited, it is difficult to generalize the safe Bayesian optimization methods based on Gaussian kernels to high dimensions. At the same time, the safe Bayesian optimization algorithm based on additive Gaussian kernels has a computing resource requirement that is too high. As mentioned in \cite{AdditiveGaussianProcesses}, when the problem dimension is high, the general standard kernel selection method is to select the first-order and highest-order additive kernels together to form a new kernel, which can take into account the high exploration range of the low-order additive kernel and the high ability of the high-order additive kernel to explore the interactions between different dimensions. However, this standard selection method has not been experimentally verified. 

In this paper, we build upon the Nyström approximate kernel selection method \cite{KernelSelection} to select the most important additive kernels based on the observation of the system (the GP prior), as shown in Algorithm \ref{algorithm}, and then to improve the additive Gaussian process-based safe Bayesian optimization algorithm. By using this approach, we achieved kernel selection with a time complexity of \(O(l)\), where \(l\) represents the number of samples in the GP prior. the Nyström approximate kernel selection method is particularly advantageous as it efficiently balances computational complexity and selection accuracy, making it well-suited for our experimental setup.

The explanations of the variables we used in Algorithm \ref{algorithm} are listed as follows:

\begin{itemize}
    \item $l$: The number of observed data points from the system (the GP prior).
    \item $\mathcal{S} = \{(x_i, y_i)\}_{i=1}^l$:
    the observed data points from the system (the GP prior), where $x_i$ represents the 9-dimensional controller parameters and $y_i$ represents the corresponding performance measurement $P$.
    \item $\mathcal{K} = \{\kappa_1, \dots, \kappa_n\}$: The set of additive kernels, which is from additive kernel 1 to additive kernel 9.
    \item $c$: The number of indices to sample from the data.
    \item $k$: A crucial parameter that determines the number of columns selected from the kernel matrix for the approximation.
    \item $\kappa$: The kernel function of the additive kernels from the set $\mathcal{K}$.
    \item $\mu$: A regularization parameter.
    \item $\mathbf{C}$: The matrix formed by the selected columns from the kernel matrix $K$.
    \item $\mathbf{W}$: A matrix derived from $C^{\top} C$.
    \item $C_{\text{ree}}(\mathbf{\tilde{K}})$: This term represents a criterion used to evaluate the approximation quality of the kernel matrix $\tilde{K}$. Specifically, it is a regularized empirical error that quantifies how well the approximate kernel $\tilde{K}$ (obtained through the Nyström method) represents the true underlying function $f_\kappa$, which is associated with the kernel $\kappa$.
\end{itemize}


According to the $C_{\text{ree}}\_\text{list}$ obtained from Algorithm \ref{algorithm}, we can determine the importance of each additive kernel based on the $C_{\text{ree}}$ values. Kernels with lower $C_{\text{ree}}$ values are considered more important and are prioritized in the selection process.

To further enhance the efficiency and computational ease of the kernel, we introduced a base-kernel selection process using a feature selection algorithm. In our experiment, we utilized a Random Forest Regressor\cite{RandomF} for feature selection due to its ability to handle high-dimensional data effectively while providing robust feature importance rankings. Its ensemble nature also helps reduce overfitting, ensuring that the selected features contribute meaningfully to the kernel selection process.

After selecting the base kernel, we generate our additive kernels by prioritizing the importance of each base kernel. We carefully organize the additive computation by adjusting the frequency with which each base kernel appears in the final model. This strategic arrangement is designed to maximize the impact of the most influential kernels, thereby improving the optimization process. By doing so, we aim to achieve a theoretically better outcome, ensuring that the optimization aligns closely with our assumptions and enhances overall performance.

By incorporating the prioritized kernels into the whole framework, we ensure that the optimization process benefits from the strategically selected kernels and the proven methods of Safe Bayesian optimization. This approach helps improve the overall effectiveness of our optimization, ensuring that the computation aligns with the theoretical improvements we've aimed for.

\begin{figure*}[t]
    \centering
    
    \vspace{-0.5cm}
    \hspace{0.5cm}
    \includegraphics[width=0.9\textwidth]{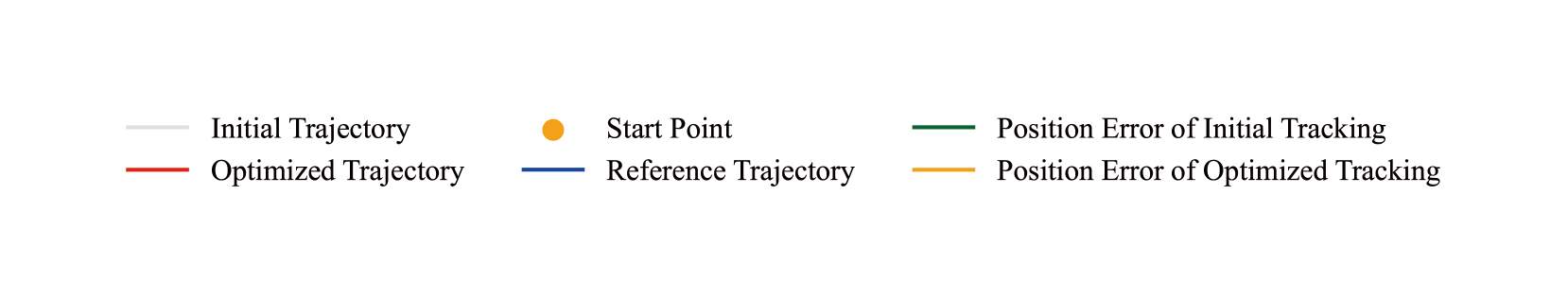}  
    \vspace{-0.7cm}  
    
    \setlength{\belowcaptionskip}{0pt}
            \begin{tabular}{ccc}
                \hspace{-0.4cm}
                \begin{subfigure}[b]{0.32\textwidth}  
                    \includegraphics[width=\textwidth]{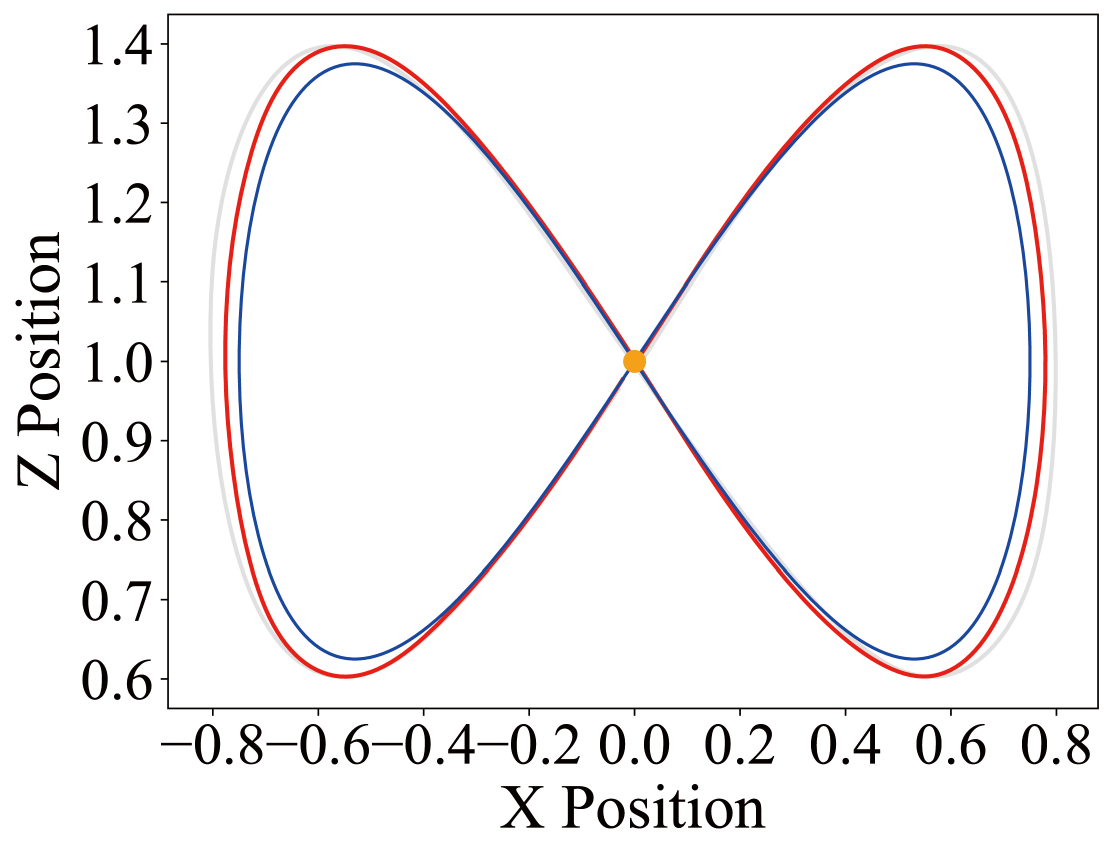}
                    \caption{\small{Trajectory tracking results.}}
                    \label{fig:compare_tracking_result}
                \end{subfigure} 
                \hspace{-0.27cm} 
                \begin{subfigure}[b]{0.32\textwidth}  
                \includegraphics[width=\textwidth]{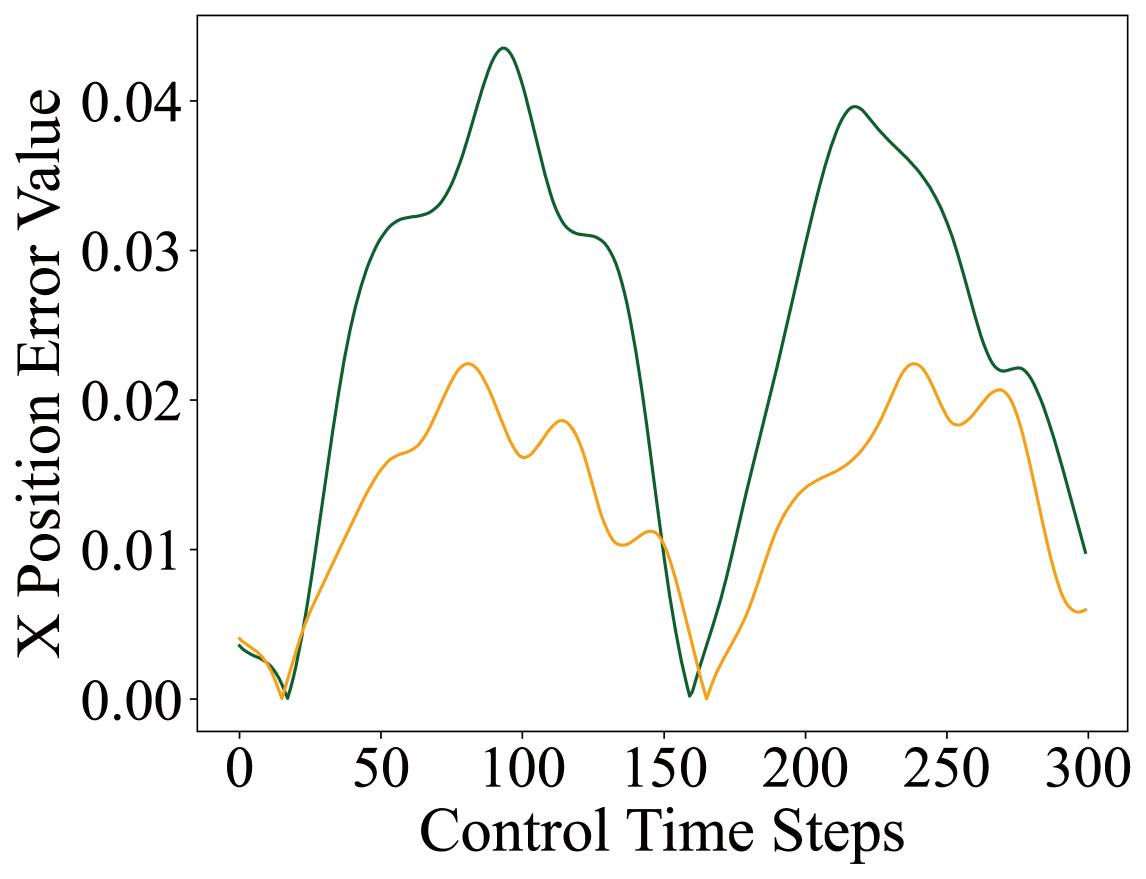}
                    \caption{\small{Comparison of $x$-axis position error.}}
                    \label{fig:compare_x_error}
                \end{subfigure} 
                \hspace{-0.27cm}
                \begin{subfigure}[b]{0.32\textwidth}  

                    \includegraphics[width=\textwidth]{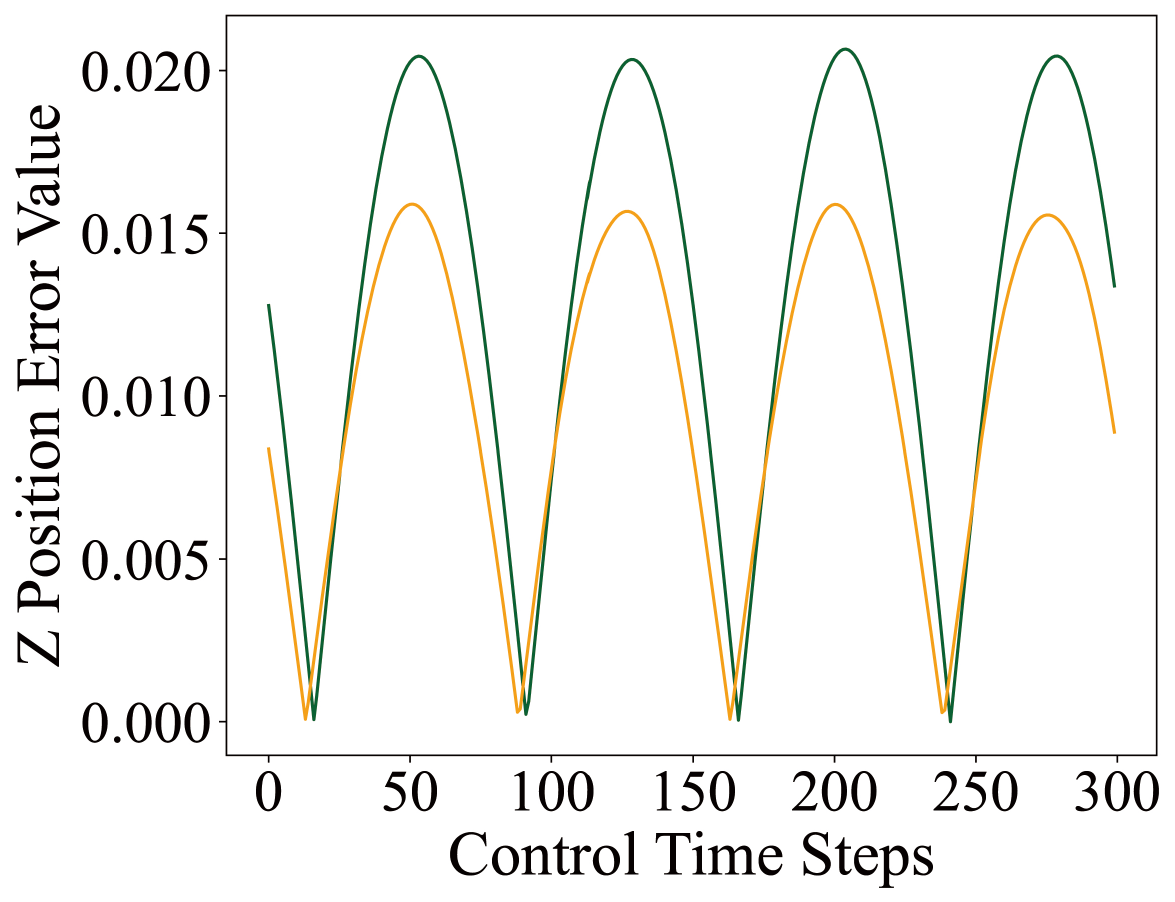}
                    \caption{\small{Comparison of $z$-axis position error.}}
                    \label{fig:compare_z_error}
                \end{subfigure} \\
            \end{tabular}
    \setlength{\belowcaptionskip}{-10pt}
    \caption{\small{Comparison of trajectory tracking results and tracking error. (a) compares the trajectory tracking curves of the initial controller and the optimal controller tuned with our method, where the blue curve is the reference trajectory, the light gray curve is the trajectory of the initial controller, the red curve is the trajectory of the optimal controller, and the yellow point is the starting point of the trajectory. (b) and (c) compare the tracking errors of the initial controller and the optimal controller on the $x$-axis and $z$-axis, respectively. The blue curve is the error of the initial controller, and the yellow curve is the error of the optimal controller.}}
    \label{fig:Result_1}
\end{figure*}

\section{Experiment}

\subsection{Setting Quadrotor Trajectory}


In the experiment, we set the reference trajectory of the quadrotor to a horizontally placed Arabic numeral $8$, as shown in the blue curve in Fig. \ref{fig:compare_tracking_result}. We set the speed and direction of the quadrotor to be randomly initialized, then sampled the trajectory from the second lap, conducted multiple experiments, and calculated the average performance and tracking error.

\subsection{Obtaining GP Prior} 



Before applying kernel selection, we use Latin Hypercube Sampling (LHS) \cite{LHS} to generate the initial observations (the GP prior). LHS is a statistical method for generating nearly random and uniform samples of parameter values from a multidimensional distribution. We generated \(\textbf{X}\), representing the 9-dimensional PID controller parameters, and its corresponding \(\textbf{Y}\), which is the value of the performance measurement \(P\). In total, we generated $36$ points as the initial observations used in the kernel selection process.

\subsection{Conducting Kernel Selection}


Using initial observations (the GP prior) generated in the previous step, we apply Algorithm \ref{algorithm} for the additive kernel selection. Based on the calculated \(C_{\text{ree}}\_\text{list}\), we select the top $3$ most important additive kernels for our optimization process. Within each selected additive kernel, we use the Random Forest Regressor for feature selection, as described in Section \ref{method:kernel_selection}. Ultimately, we select all the essential kernels to construct the reduced additive kernel for optimization.

\subsection{Optimizing Controller Parameters}

In our approach, we specifically focus on tuning the PID controller parameters to optimize the quadrotor's trajectory tracking performance. By adjusting these parameters, we aim to enhance the control system’s ability to follow a reference trajectory accurately.

In this section, we implement our method for controller tuning alongside several baseline approaches for comparison, i.e., the standard kernel selection introduced in \cite{AdditiveGaussianProcesses}, LINEBO, and unconstrained BO. These baselines provide a benchmark to evaluate the effectiveness of our proposed approach, allowing us to assess how well our method performs relative to established techniques.


\section{Results and Analysis}

\subsection{Kernel Selection Results}


Given that Algorithm \ref{algorithm} contains random settings, we conducted the computation $1000$ times to get an average estimation of each kernel's contribution, enhancing the robustness and reliability of the results. Table \ref{table} illustrates the average \(C_\text{ree}\) for each additive kernel, calculated from the GP prior.
\begin{table}[b]
\setlength{\abovecaptionskip}{0pt}
\setlength{\belowcaptionskip}{-2pt}
\caption{\small{Average \(C_\text{ree}\) of additive kernels in different orders. A smaller \(C_{\text{ree}}\) value indicates a better-performing kernel.}}
\label{table}
\begin{center}
\begin{tabular}{c c}
\toprule
\textbf{Order of Additive Kernels} & \textbf{Average \(C_\text{ree}\) after 1000 Trials}\\
\midrule
\textbf{1} & 22084908.8692\\
\midrule
\textbf{2} & 21002011.4674\\
\midrule
\textbf{3} & 21123139.7816\\
\midrule
\textbf{4} & 21358712.5683\\
\midrule
\textbf{5} & 21470536.6469\\
\midrule
\textbf{6} & 21626907.8987\\
\midrule
\textbf{7} & 21887200.5901\\
\midrule
\textbf{8} & 22012720.6549\\
\midrule
\textbf{9} & 23492448.2242\\
\bottomrule
\end{tabular}
\end{center}
\end{table}

\begin{figure*}[t]
    \centering
    
    \vspace{-0.3cm}
    \hspace{0.5cm}
    \includegraphics[width=0.85\textwidth]{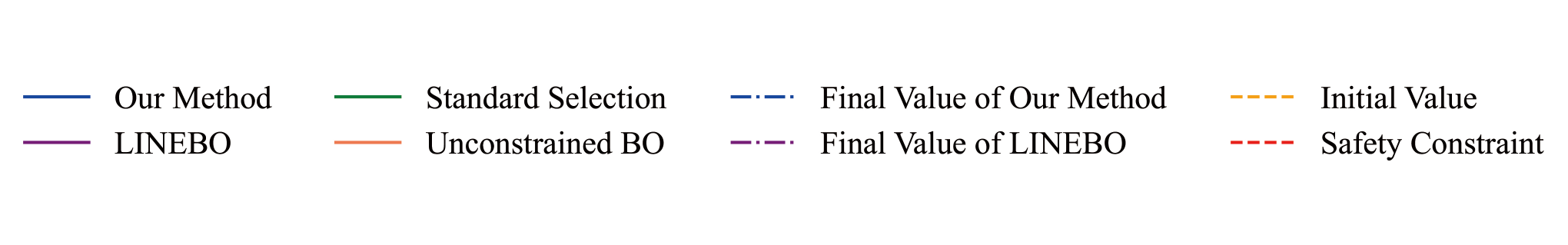}  
    \vspace{-0.5cm}  
    
    \begin{tikzpicture}
    \setlength{\belowcaptionskip}{0pt}
        \node[anchor=south west,inner sep=0] (image) at (0,0) {
            \begin{tabular}{ccc}
                \hspace{-0.4cm}
                \begin{subfigure}[b]{0.32\textwidth}  
                    \includegraphics[width=\textwidth]{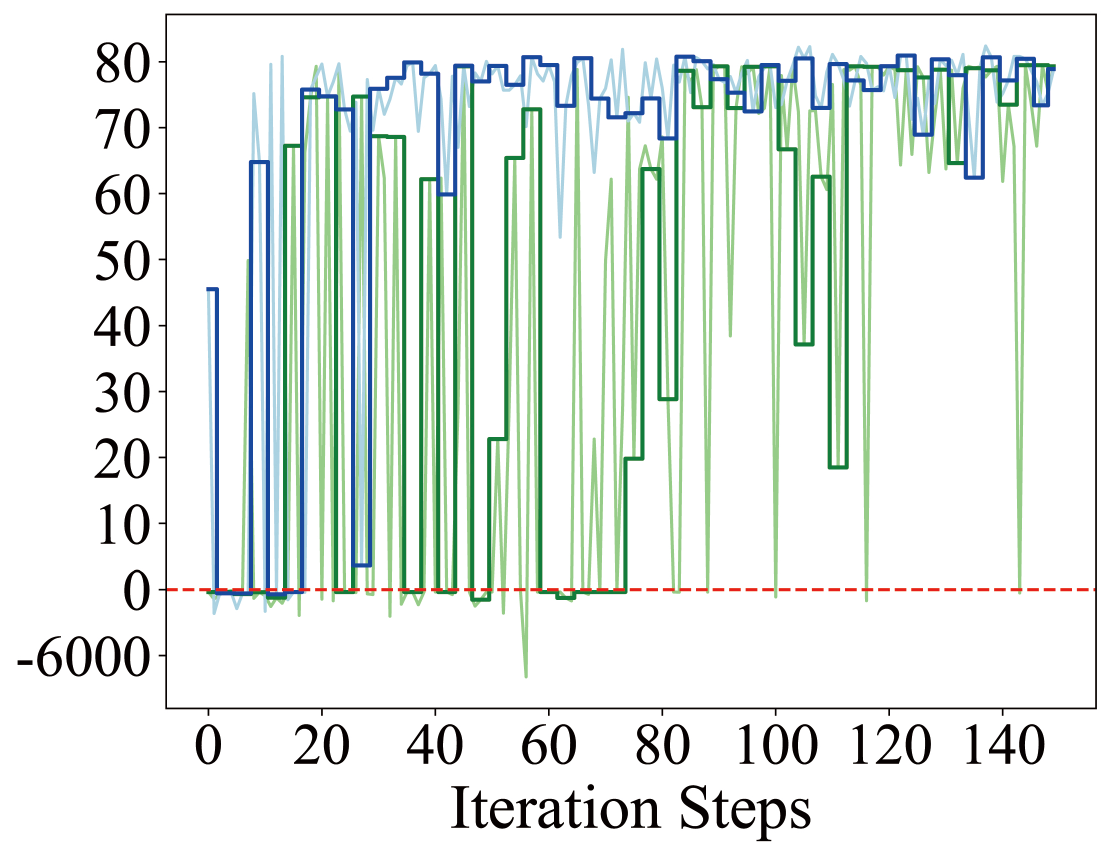}
                    \caption{\small{Compare with standard kernel selection.}}
                    \label{fig:compare_a}
                \end{subfigure} 
                \hspace{-0.27cm} 
                \begin{subfigure}[b]{0.32\textwidth}  
                \includegraphics[width=\textwidth]{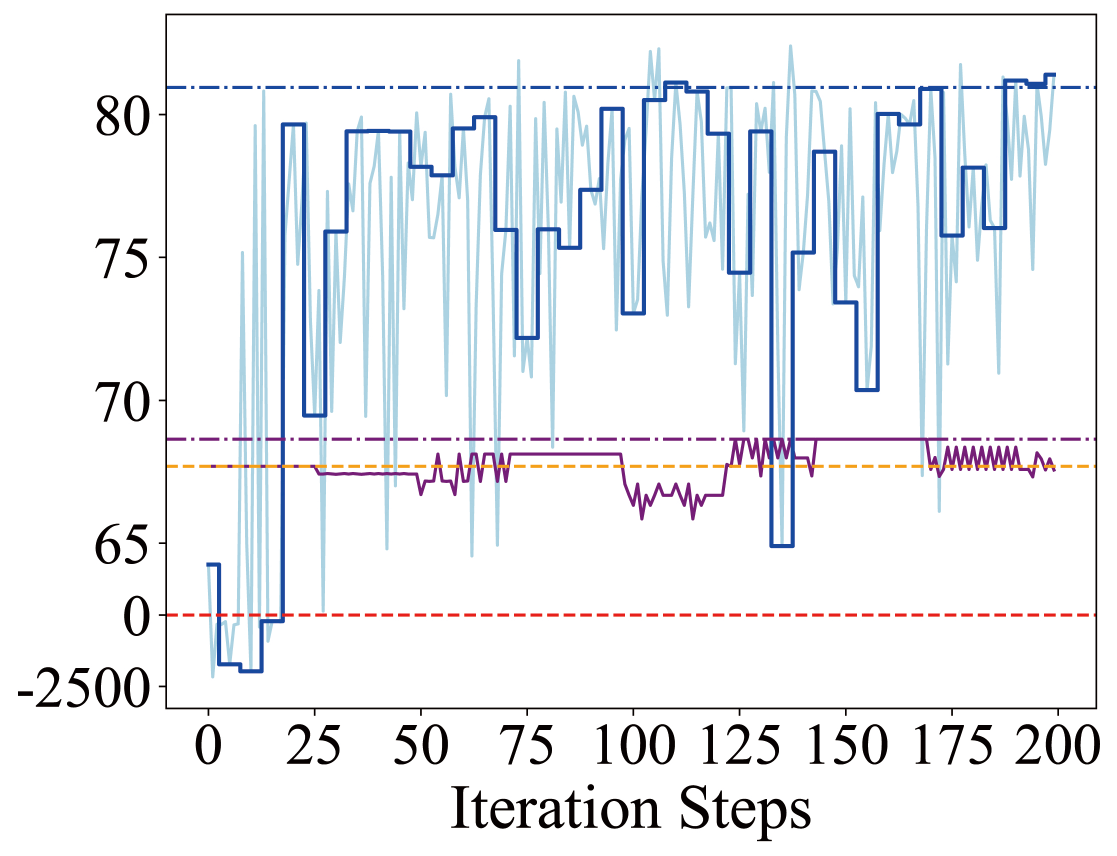}
                    \caption{\small{Compare with LINEBO.}}
                    \label{fig:compare_b}
                \end{subfigure} 
                \hspace{-0.27cm}
                \begin{subfigure}[b]{0.32\textwidth}  

                    \includegraphics[width=\textwidth]{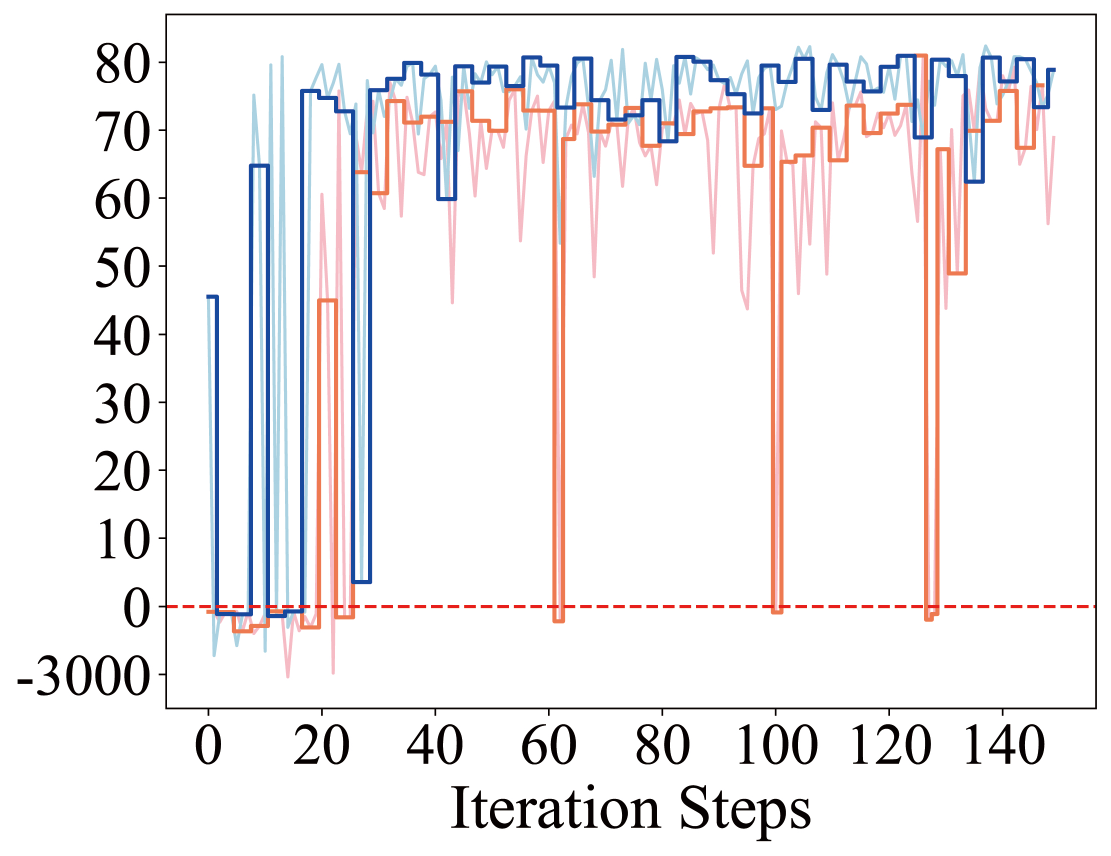}
                    \caption{\small{Compare with Unconstrained BO.}}
                    \label{fig:compare_c}
                \end{subfigure} \\
            \end{tabular}
        };
        \node[rotate=90, anchor=south] at ([yshift=0.7cm,font=\small]image.west) {Performance Value};
    \end{tikzpicture}
    \setlength{\belowcaptionskip}{-10pt}
    \caption{\small{Comparison of optimization performance between our method and different baseline methods. (a) compares our method with the standard kernel selection method introduced in \cite{AdditiveGaussianProcesses}, (b) compares our method with the state-of-the-art high-dimensional safe Bayesian optimization method LINEBO, (c) compares our method with unconstrained Bayesian optimization method.}}
    \label{fig:Comparison with baselines}
\end{figure*}

\subsection{Control Optimization Results}


Based on the results in Table \ref{table}, we identified the three most important additive kernels as order $2$, $3$, and $4$. By applying these selected kernels to the additive Gaussian process-based safe Bayesian optimization algorithm, we obtained the following trajectory tracking results.






Fig. \ref{fig:Result_1} compares the trajectory results between the initial controller and the optimal controller tuned with our method and the tracking errors in the $x$ and $z$-axis. By comparing the initial trajectory and the optimized trajectory visually, it is shown that the optimized trajectory aligns more closely with the reference trajectory and has a smaller position error than the initial trajectory.

Our optimization process significantly improved the tracking performance, as reflected by the performance reward \(J^R\), which increased from $-62.7015$ to $-35.8925$, representing an improvement of $42.75\%$. As a result, this enhancement underscores the effectiveness of our algorithm in fine-tuning the control parameters to achieve better trajectory tracking accuracy.

\subsection{Comparison to Baseline Methods}

\subsubsection{Using different Kernels}

In Fig. \ref{fig:compare_a}, we compare our selected additive kernels $2$, $3$, and $4$ against the standard selection method that combines additive kernels $1$ and $9$. We conducted a total of 150 iterations and obtained the results. Our method stabilizes around the optimum, about $81$, after approximately 80 iterations, exhibiting minimal oscillation.

In contrast, the result achieved using the standard selection method fails to exceed a value of $80$ in terms of performance measurement across all iterations. This demonstrates that our kernel selection method is more effective in achieving higher performance and stability than the baseline approach.


\subsubsection{Comparison to LINEBO}


In Fig. \ref{fig:compare_b}, we compare our method with LINEBO by conducting the optimization process over 200 iterations. While LINEBO achieved a performance value of $68.6420$ within 200 iterations, our method converges more quickly, reaching a final optimal value of $81.0591$ after approximately 80 iterations. This result demonstrates that our method is more efficient than LINEBO. Although LINEBO might eventually reach a similar optimal value after a thousand iterations, our approach appears to be faster and less computationally intensive.

\subsubsection{Comparison to Unconstrained BO}


In Fig. \ref{fig:compare_c}, we compare our method to Unconstrained BO. The figure shows that after 20 iterations, our method avoids evaluating any unsafe values, while Unconstrained BO violates the safety constraint several times. Due to effective kernel selection, unconstrained BO also converges to a good optimal value, which is slightly lower than ours. However, in real-world scenarios, breaking the safety constraint could lead to critical failures, such as damaging a quadrotor. Therefore, our method is not only more reliable but also safer for real-world applications.

\section{Conclusions}


We enhanced Safe Bayesian Optimization via Additive Gaussian Processes by incorporating a kernel selection algorithm, resulting in a more efficient and computationally less demanding method. This improved approach is tested on a quadrotor trajectory tracking task in Safe Control Gym by tuning the 9-dimensional PID controller parameters, and compared with baseline methods such as the widely used standard kernel selection scheme, the state-of-the-art high-dimensional safe Bayesian optimization method LINEBO, and the unconstrained Bayesian optimization method. The experimental results demonstrate that our method achieves an ideal tracking result, improving the performance measurement by $42.75\%$, and outperforms the baseline methods in terms of tracking accuracy and tuning efficiency.

\addtolength{\textheight}{-6.5cm}   








\bibliographystyle{ieeetr}
\bibliography{references}

\end{document}